%% file: main.tex
\documentclass[letterpaper, 10 pt, conference]{ieeeconf}

\IEEEoverridecommandlockouts                              
\usepackage{graphics} 
\usepackage{epsfig} 
\usepackage{subfigure}
\usepackage{graphicx}
\usepackage{multirow}
\usepackage{arydshln}
\usepackage{xcolor}
\usepackage{cite}

\title{	\LARGE \bf Implicit LiDAR Network: \\LiDAR Super-Resolution via Interpolation Weight Prediction }

\author{Youngsun Kwon$^{1}$, Minhyuk Sung$^{2}$* and Sung-Eui Yoon$^{3}$*
	\thanks{Y. Kwon, M. Sung and S. Yoon are with School of Computing, Korea Advanced Institute of Science and Technology, Daejeon, South Korea. M. Sung and S. Yoon are co-corresponding authors.
		{\tt\small youngsun.kwon@kaist.ac.kr, \tt\small mhsung@kaist.ac.kr, \tt\small sungeui@kaist.edu}
	}%
}

\begin{document}
\maketitle
\thispagestyle{empty}
\pagestyle{empty}

\input{0_abstract} 
\input{1_introduction}
\input{2_approach}
\input{3_result}
\input{4_conclusion}

\bibliographystyle{ieee}
\bibliography{references}

\end{document}

%% file: 0_abstract.tex
\begin{abstract}

Super-resolution of LiDAR range images is crucial to improving many downstream tasks such as object detection, recognition, and tracking. 
While deep learning has made a remarkable advances in super-resolution techniques, typical convolutional architectures limit upscaling factors to specific output resolutions in training. 
Recent work has shown that a continuous representation of an image and learning its \emph{implicit} function enable almost limitless upscaling. 
However, the detailed approach, predicting \emph{values (depths)} for neighbor pixels in the input and then linearly interpolating them, does not best fit the LiDAR range images 
since it does not \emph{fill} the unmeasured details but \emph{creates} a new image with regression in a high-dimensional space. 
In addition, the linear interpolation blurs sharp edges providing important boundary information of objects in 3-D points. 
To handle these problems, we propose a novel network, Implicit LiDAR Network (ILN), which learns not the values per pixels but \emph{weights} in the interpolation 
so that the super-resolution can be done by blending the input pixel depths but with non-linear weights. 
Also, the weights can be considered as \emph{attentions} from the query to the neighbor pixels, and thus an attention module in the recent Transformer architecture can be leveraged. 
Our experiments with a novel large-scale synthetic dataset demonstrate that the proposed network reconstructs more accurately than the state-of-the-art methods, achieving much faster convergence in training. 

\end{abstract}

%% file: 1_introduction.tex
\section{Introduction}
\label{sec:introduction}

LiDAR sensor capturing 3-D geometry of the surrounding environment is essential in many intelligent systems such as autonomous vehicles and service robots, 
enabling various robotics/vision tasks such as object detection, recognition, tracking, and motion planning.
The performance of such tasks is often susceptible to the density of the sensed point cloud, i.e., the resolution of the LiDAR range image.
However, increasing the sensing resolution takes longer capture time, much more energy consumption, and higher cost. 
Hence, super-resolution techniques generating a higher resolution image from a lower resolution one have been actively studied and applied to LiDAR scan data. 

The advance of deep learning has made significant improvements in super-resolution techniques. 
The typical approach in previous work~\cite{dong2015image, kim2016accurate, zhang2018residual, shan2020simulation} is to build an autoencoder-style architecture based on convolution/deconvolution layers. 
Although this approach recovers fine details well, its network is constrained to produce the output image with a specific target resolution used in training and thus restricts its applicability to diverse systems. 
In light of the recent success of learning an implicit function for 3-D shape reconstruction~\cite{mescheder2019occupancy, peng2020convolutional, park2019deepsdf, mildenhall2020nerf, rist2021semantic}, 
Chen et al.~\cite{chen2021learning} first proposed to view an image as \emph{continuous} 2-D data and predict an implicit function that returns color for a given query point. 
Their method, called Local Implicit Image Function (LIIF), showed how a super-resolution network could be trained without specifying the output resolution, 
while even achieving better results than the autoencoder-style networks.

\begin{figure}[t]
	\centering
	\subfigure [LiDAR super-resolution results using LIIF~\cite{chen2021learning} and ours.]
	{ \includegraphics[width=8.2cm]{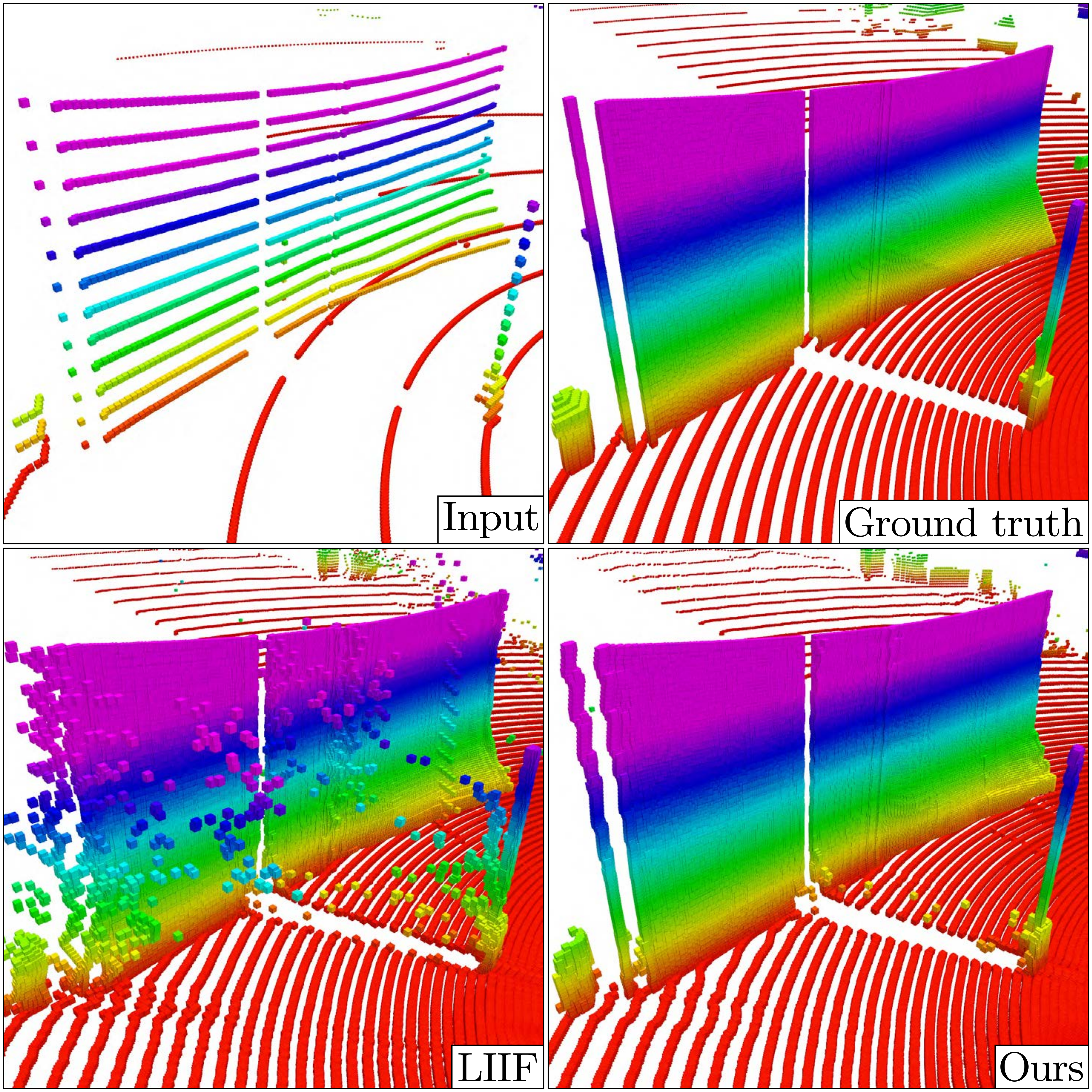} }
	\subfigure [Architecture comparison between LIIF~\cite{chen2021learning} and ours.]
	{ \includegraphics[width=8.2cm]{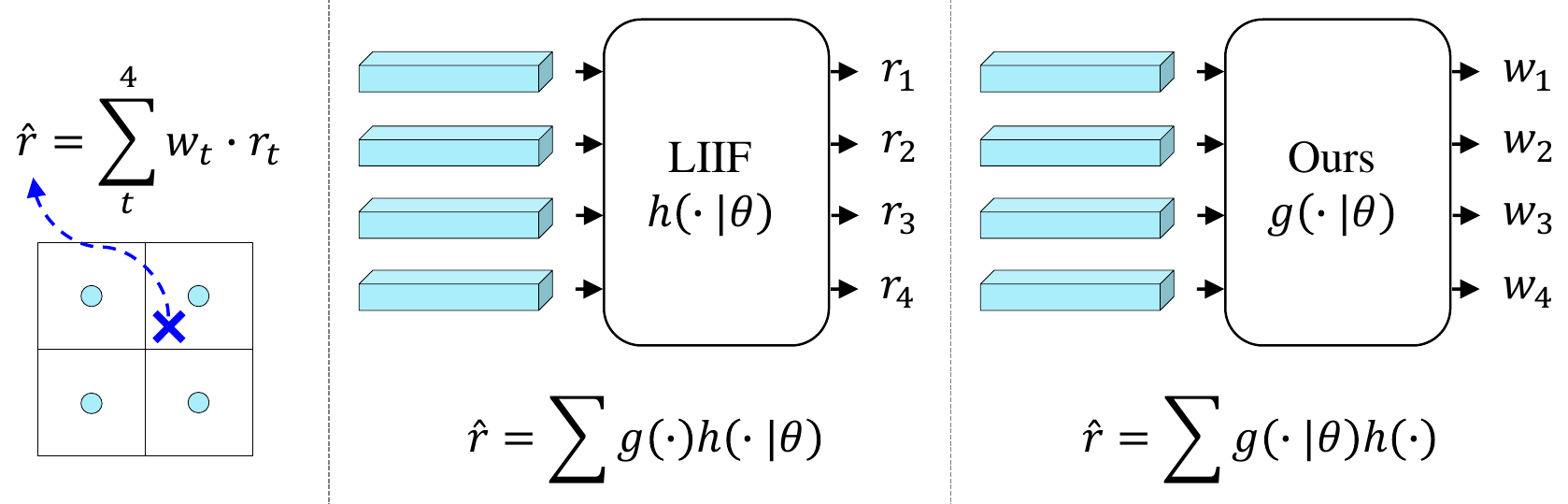} }
	\caption{
	    Super-resolution results and network architecture comparison.
	    $(a)$ shows the reconstruction of dense LiDAR points using the sparse input, where color represents relative elevation of structures.
	    $(b)$ summarizes the difference between two implicit networks - value prediction (LIIF) and weight prediction (Ours), where $\theta$ indicates the learning parameters of network. 
	    Our method predicts the interpolation weight $w_t$ instead of depth value $r_t$, resulting in the robust super-resolution $(a)$.
	}
	\label{fig:sharp_transition}
\end{figure}

Although LIIF can be directly applied to upscale LiDAR, the detailed idea has some problems, particularly when used for the LiDAR range images. 
LIIF represents the implicit function as a linear interpolation of neighbor pixels, but instead of the pixel values (depths in our case) in the input, \emph{predicted} depth values are used. 
Thus, it turns out that the network does not \emph{fill} the missing detailed information in the super-resolution but \emph{creates} a new image looking the same as the input; it makes the training very time-consuming.
Also, the problem becomes a regression problem in a very high-dimensional space (the dimension of the number of pixels), which also adds more difficulty in network training. 
Moreover, while the depths for the input image pixels are learned, still the depths are \emph{linearly} interpolated. 
It means that \emph{sharp edges} (depth value transitions) are prone to be blurred, as shown in Fig.~\ref{fig:sharp_transition}-$(a)$. 
For LiDAR range images, it is crucial to precisely reconstruct the sharp edges, since small errors in the image pixels can result in a significant difference in the 3-D space, 
largely affecting the performance in downstream applications. 

To handle these problems, we propose a novel network, Implicit LiDAR Network (ILN), which stems from the idea of learning implicit function for super-resolution. 
In contrast to LIIF, our ILN does not predict the depths of input image pixels but the \emph{weights} for the interpolation; see the difference in Fig.~\ref{fig:sharp_transition}-$(b)$. 
This change makes a big difference in the network training since it does not learn how to make a \emph{new} image but how to blend the pixel values to \emph{fill} the fine details. 
Furthermore, such network design makes the training to be converged much faster.
In our model, the weights for each query to the neighbor pixels can also be viewed as \emph{attentions}, 
and thus a recent attention module such as one in Transformer~\cite{dosovitskiy2020image} can be leveraged to achieve the best performance. 
Most importantly, sharp edges can be reconstructed more accurately as shown in Fig.~\ref{fig:sharp_transition}-$(a)$ since the interpolation is no longer linear.

To this end, we introduce an architecture predicting the weights for interpolation based on the Transformer~\cite{dosovitskiy2020image} self-attention module 
and then conduct experiments by training networks with a novel synthetic large-scale dataset created using CARLA simulator~\cite{Dosovitskiy17} (LiDAR scanning in virtual outdoor scenes).
We compare our method with three baselines, bilinear interpolation, LIDAR-SR (the most recent autoencoding-style network)~\cite{shan2020simulation}, and LIIF~\cite{chen2021learning}, 
and show that our method achieves the best performance. 

In summary, our contributions are the followings:
\begin{itemize}
\item We propose a novel network, Implicit LiDAR Network (ILN), predicting an implicit function based on neighbor pixel interpolation for limitless LiDAR super-resolution.
\item In the neighbor pixel interpolation, we demonstrate that learning weights is more effective than learning pixel values (depths) as adopted by a prior work~\cite{chen2021learning}.
\item We propose to use an attention module of Transformer~\cite{dosovitskiy2020image} by viewing the weights as attentions.
\item We introduce a novel large-scale synthesis benchmark for LiDAR super-resolution created using CARLA simulator~\cite{Dosovitskiy17}, and show the outperforming results of our method in the experiments compared with baselines.
\end{itemize}

%% file: 2_approach.tex
\begin{figure*}[ht]
	\centering
	\includegraphics[width=\linewidth]{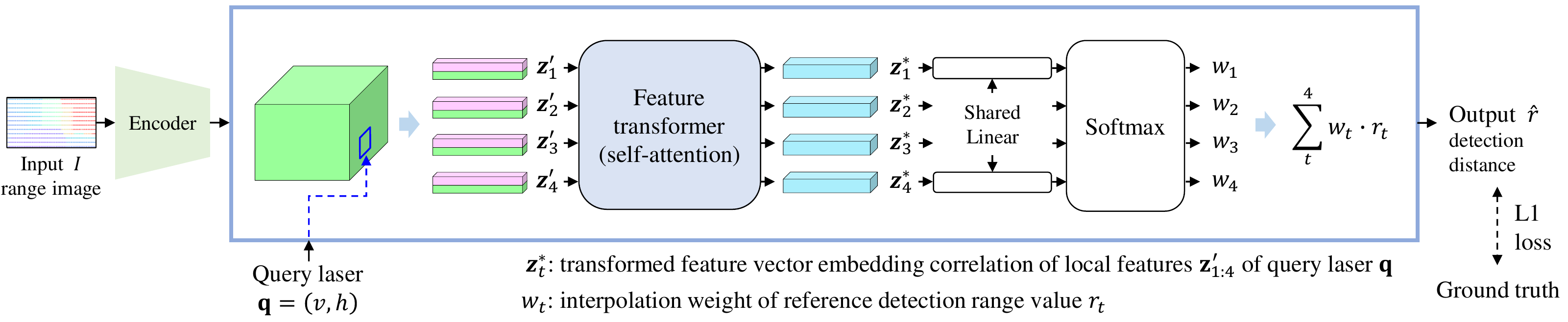}
	\vspace{-0.7cm}
	\caption{Framework of Implicit LiDAR Network (ILN). 
		The proposed model predicts the interpolation weights $w_{1:4}$ with local deep features $\mathbf{z}'_{1:4}$ of the query laser $\mathbf{q}$. 
		Noticeably, the self-attention module enables the accurate detection range prediction $\hat{r}$ of the query laser. 
		See Fig.~\ref{fig:sub_structures} for more details.
	}
	\label{fig:framework}
\end{figure*}

\begin{figure}[ht]
	\centering
	\subfigure [Embedding query information to features] 		{ \includegraphics[width=3.1in]{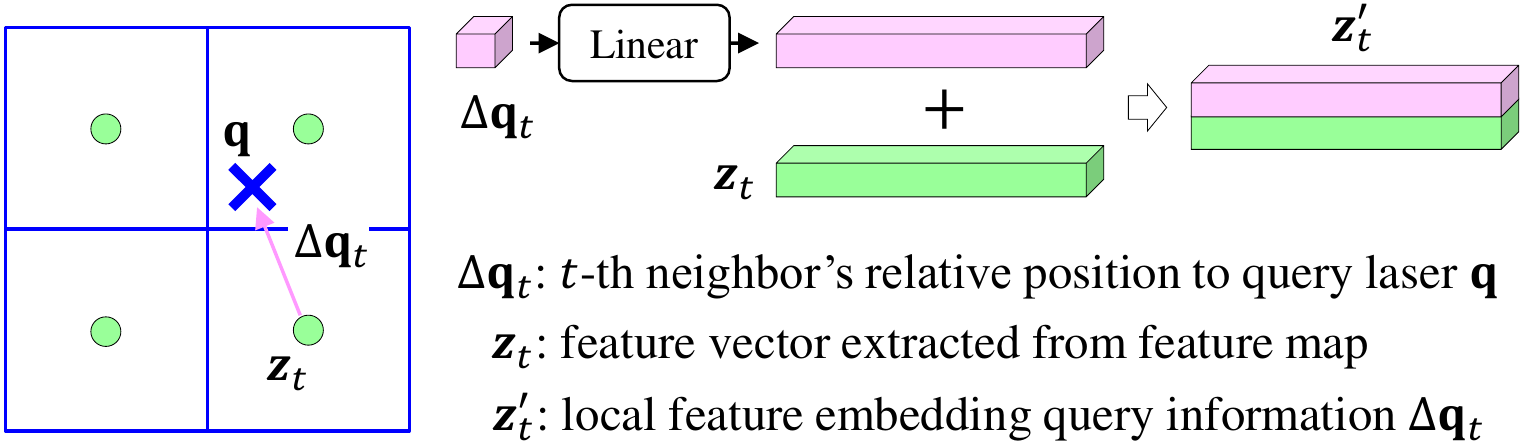} }
	\subfigure [Self-attention module in feature transformer]   { \includegraphics[width=3.1in]{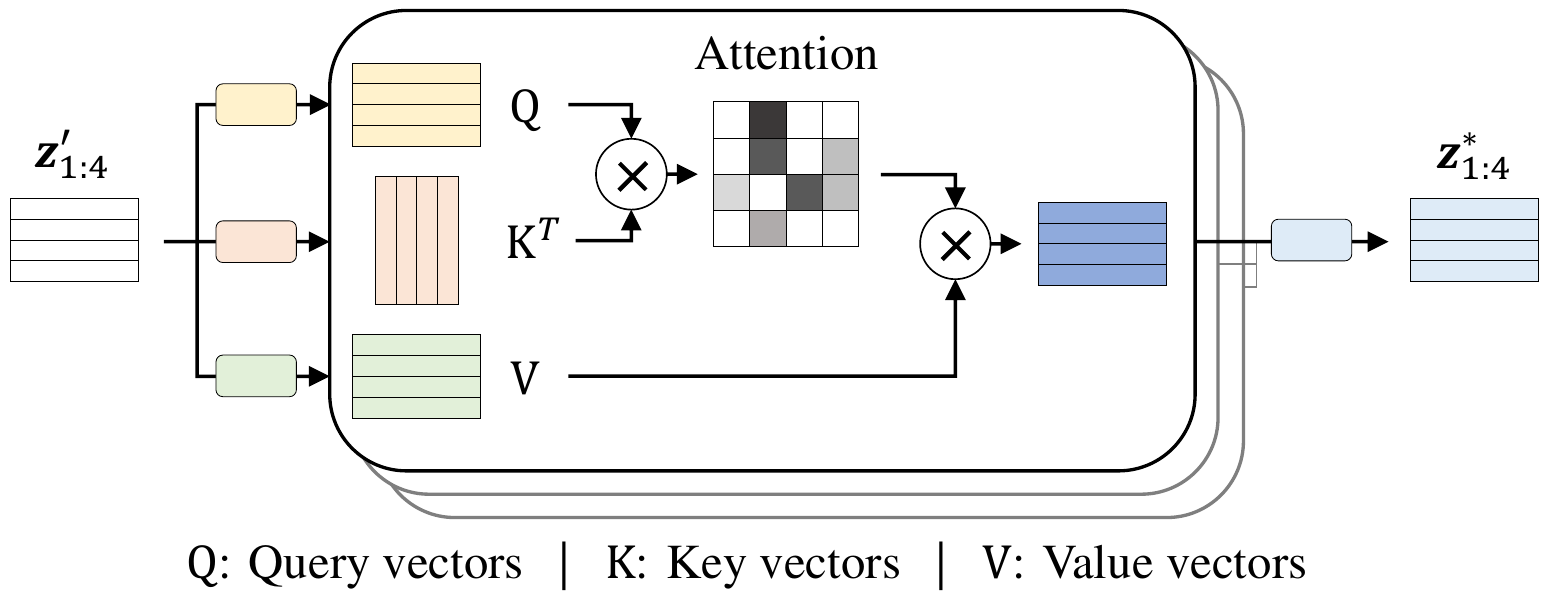} }
	\caption{Local query embedding and self-attention module.
		$(a)$ represents a process embedding the query information, relative position $\Delta \mathbf{q}_t$, to compose a neighbor's feature vector $\mathbf{z}_{t}'$.
		$(b)$ shows the self-attention module that extracts the correlation of local features $\mathbf{z}_{1:4}'$ as an attention map 
		and produces the correlated features $\mathbf{z}_{1:4}^{*}$ for robust detection range prediction.
    }
	\vspace{-0.3cm}
	\label{fig:sub_structures}
\end{figure}

\section{Implicit LiDAR Network}
\label{sec:approach}

\subsection{Problem Definition and Motivation}
\label{sec:overview}

A range-based sensor shoots multiple lasers and measures the depth (detection distance) of each laser. 
Let $v$ and $h$ be vertical and horizontal directions of a laser, and $r$ be its measurement depth value. 
Then, we can represent the measurement points in the sensor coordinate as a set of 2-D depth samples, where each sample indicates the depth $r$ of the laser $(v, h)$. 
Based on this sensor model, the sample set can be represented as a range image since a real LiDAR has a sensing resolution.
In the range image, each pixel indicates the depth $r$ at the pixel center $(v, h)$.
 
The goal of resolution-free LiDAR is to predict a detection distance $\hat{r}$ of a query laser $\mathbf{q}$ based on an input range image $I$. 
Then, our problem becomes finding an unknown function $f(\cdot)$ expressed as
\begin{equation}
\label{eq:upsampling}
\hat{r}=f(I, \mathbf{q}),
\end{equation}
where $\mathbf{q}$ means the query laser's direction $(v, h)$ within the sensor's field of view.
The state-of-the-art method, LIIF~\cite{chen2021learning}, solves this problem as:
\begin{equation}
\label{eq:liif}
\hat{r}=\sum_{t}^4 g(\cdot)h(\cdot|\theta)=\sum_{t}^4 \frac{S_t}{S} \cdot h(\mathbf{z}_{t}'|\theta),
\end{equation}
where $g(\cdot)$ and $h(\cdot)$ denote the weight and value functions, respectively. 
The network $h(\mathbf{z}_{t}'|\theta)$ predicts the value of query's $t$-th neighbor pixel by using its local feature $\mathbf{z}_{t}'$ and the learning parameters $\theta$, 
while computing each weight $S_t$ based on the distance to the query.
The problem of this approach is that the network learns new values (depths in our case) for input pixels instead of using \emph{given} values, 
and thus the output can largely deviate from the input in the early stage of training. 
Moreover, LiDAR range images typically have lots of sharp edges, while the edges may not be reconstructed well with linear interpolation (Fig.~\ref{fig:sharp_transition}-$(a)$).

To overcome these problems, this work proposes a novel approach, named Implicit LiDAR Network (ILN), predicting the weights:
\begin{equation}
\label{eq:iln}
\hat{r}=\sum_{t}^4 g(\cdot|\theta)h(\cdot)=\sum_{t}^4 g(\mathbf{z}_{t}'|\theta) \cdot r_t.
\end{equation}
Our model utilizes the neighbor pixel value $r_t$ of input image instead of the prediction value. 
The proposed network $g(\mathbf{z}_{t}'|\theta)$ predicts the interpolation weight with the deep feature embedding prior knowledge.
The predicted weight determines which neighbors are valuable to infer the detection distance $\hat{r}$ of query laser $\mathbf{q}$.
This approach focuses on how to fill the unmeasured information with the neighbor pixels (sensor observations), resulting in fast convergence speed and robust LiDAR points reconstruction as well. 
In the following section, we introduce technical details of the proposed structure (Fig.~\ref{fig:framework}) to predict interpolation weights.

\subsection{Interpolation weight estimation}
\label{sec:weight_estimation}

\textbf{Local feature extraction.}
An input range image $I$ consists of the sensor observations, i.e., detection distances of lasers. 
Nonetheless, its individual pixel has insufficient information to predict interpolation weights robustly.
We, therefore, extract deep features from the input low-resolution range image.
The pixel-based representation of the input enables to utilize the well-studied feature extractors~\cite{simonyan2014very, he2016deep, Lim_2017_CVPR_Workshops}. 
In this work, we opt the feature encoder~\cite{Lim_2017_CVPR_Workshops} that the state-of-the-art method~\cite{chen2021learning} uses. 
The encoder module captures local contexts of pixels through deep convolutional operations, and represents the input range image as a feature map. 

To predict a detection distance $\hat{r}$ of a query laser $\mathbf{q}$, our implicit model utilizes the deep features $\mathbf{z}_{1:4}$ located in the query's neighbor pixels of the feature map.
However, since each feature vector has no query information for the detection range prediction, we need to embed such information into the neighbor's feature.
As shown in Fig~\ref{fig:sub_structures}-$(a)$, our model uses a relative position $\Delta \mathbf{q}_t$ between a pixel center 
and the query point to generate a local feature vector $\mathbf{z}_{t}'$, similar to the local implicit model~\cite{chen2021learning}. 
On the other hand, unlike the prior work, our network adopts the positional embedding on feature space in the light of their great success~\cite{vaswani2017attention, dosovitskiy2020image, mildenhall2020nerf}.

\textbf{Feature transformation using self-attention.}
Each local feature $\mathbf{z}_{t}'$ can be used directly to predict its interpolation weight $w_{t}$ of pixel value $r_t$.
However, we found that four predicted weights $w_{1:4}$ determine which reference value $r_t$ should be focused on for the robust final prediction $\hat{r}$.
Based on this observation, we consider the weights as \emph{attentions} from each query to its neighbor pixels, and thus leverage an attention mechanism to achieve performance improvement.

Our model (Fig.~\ref{fig:framework}) applies a self-attention module to the local features $\mathbf{z}_{1:4}'$ having the query and its neighbor pixels' information.
We found that the self-attention have achieved outstanding performance in Transformer models~\cite{vaswani2017attention, dosovitskiy2020image} on natural language processing and vision tasks as well.
In the light of the achievement, our method uses the self-attention mechanism of the recent model~\cite{dosovitskiy2020image}.

In a high-level idea, the self-attention of the feature transformer fuses the information of local features $\mathbf{z}_{1:4}'$ so that the predicted weights $w_{1:4}$ determine the reference values $r_{1:4}$ reasonably.
Fig.~\ref{fig:sub_structures}-$(b)$ shows a self-attention process where an attention map represents correlation among the local features.
The Q and K vector sets, originated from the local features, extract the self-correlation that can lead to a good choice for interpolation. 
Then, this module combines the extracted attention map and the transformed V vectors, resulting in the correlated features $\mathbf{z}_{1:4}^{*}$.
In the training step, the transformer learns its parameters to catch the best correlation of input features and thus predict the detection distance $\hat{r}$ accurately (Sec.~\ref{sec:ablation_study}).

\textbf{Interpolation weight prediction.}
As shown in Fig.~\ref{fig:framework}, the shared linear layer projects the output features $\mathbf{z}_{1:4}^{*}$ of the transformer into weight scores, 
and then the softmax function  computes the interpolation weights $w_{1:4}$. 
At the final stage, we apply Eq.~\ref{eq:iln} to infer the detection distance $\hat{r}$ of query laser $\mathbf{q}$ by combining the reference values $r_{1:4}$ and the predicted weights $w_{1:4}$.
Our method utilizes the interpolation values $r_{1:4}$ from the input range image $I$, while the recent approach~\cite{chen2021learning} predicts the values via a deep network (Eq.~\ref{eq:liif}).
In the LiDAR super-resolution, we observed that the frequent sharp transitions of range values could make unstable predictions of missing information, and thus result in lots of undesired artifacts (Fig.~\ref{fig:qualitative}). 
Under this observation, our network focuses on learning the adaptive weights prediction through deep prior knowledge, instead of the values prediction.

%% file: 3_result.tex
\section{Experimental Results}
\label{sec:result}

\subsection{Dataset for resolution-free LiDAR}
\label{sec:dataset_preparation}

An implicit network predicts an output signal at any continuous query point.
Hence, it could be the best solution for training the network with detection distance samples from infinite resolution LiDAR.
Unfortunately, there is no sensor having such hardware specification in the real world, thus it is challenging to prepare the dataset. 
In this paper, we use the CARLA simulator~\cite{Dosovitskiy17} to overcome this problem.
The simulator supports ray-based sensing simulation in various realistic environments.

We can measure ground truth detection distances at various resolution settings in the simulation environments.
Furthermore, ideally, it can be possible to train an implicit network while obtaining range samples at any continuous laser direction.
However, such online sampling and learning need intractable training time as well as computational resources. 
This paper avoids this practical issue by simulating extremely high-resolution LiDAR and collecting tremendous detection range samples.
We gather the LiDAR data at the maximum 256 and 4096 for vertical and horizontal resolutions, respectively, in which our computation resources are available.

\begin{figure}[t]
	\centering
	\subfigure [Simulation scene]   { \frame{\includegraphics[trim={0 1.8cm 0 0}, clip, width=4.0cm]{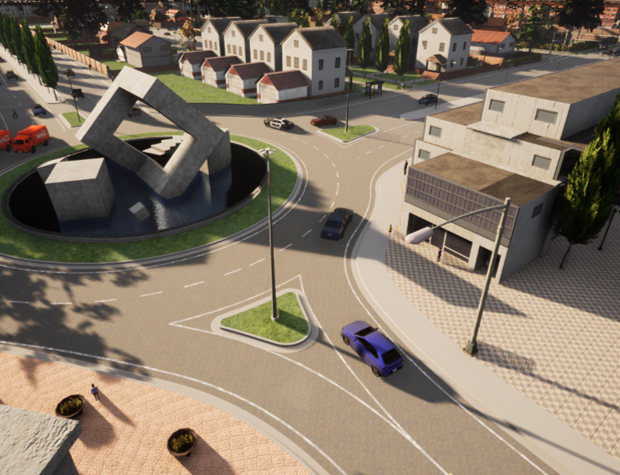}} }
	\subfigure [16 $\times$ 1024] 	{ \frame{\includegraphics[trim={0 1.8cm 0 0}, clip, width=4.0cm]{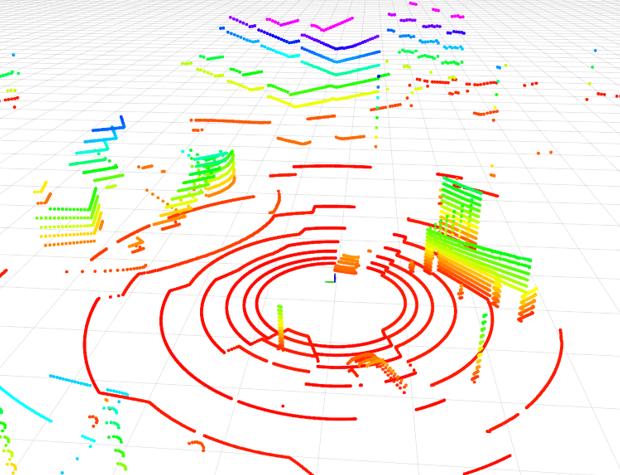}} }
	\subfigure [64 $\times$ 1024]   { \frame{\includegraphics[trim={0 1.8cm 0 0}, clip, width=4.0cm]{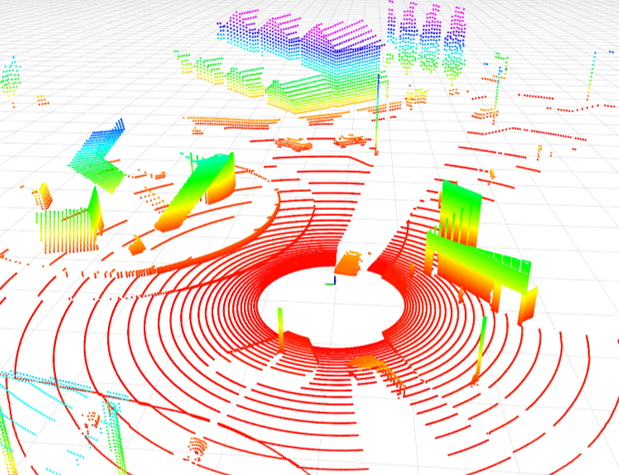}} }
	\subfigure [256 $\times$ 4096]	{ \frame{\includegraphics[trim={0 1.8cm 0 0}, clip, width=4.0cm]{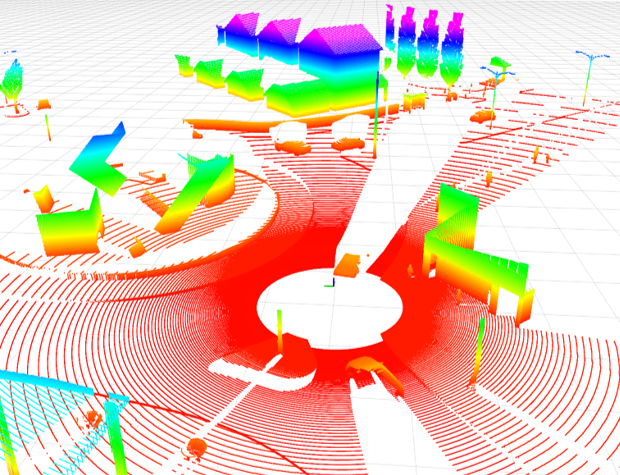}} }
	\caption{
		CARLA dataset. We simulate LiDARs having different resolutions at about 24~K waypoints of a vehicle in 8 scenes.
		The labels of $(b)-(d)$ indicate the vertical and horizontal resolutions of LiDAR, respectively. 
	}
	\vspace{-0.2cm}
	\label{fig:carla_dataset}
\end{figure}

As shown in Fig.~\ref{fig:carla_dataset}, we prepare the LiDAR data with four different resolutions; $16 \times 1024$, $64 \times 1024$, $128 \times 2048$, and $256 \times 4096$.
Such multi-resolution settings enable to train the implicit as well as pixel-based super-resolution approaches.
In addition, we can measure the super-resolution performances of implicit networks in the various test resolutions.
Table~\ref{tab:datasets} denotes the summary of our dataset configurations.

\begin{table}[t]
	\caption{Summary of CARLA dataset configurations: \\LiDAR specification and scene split.}
	\renewcommand \arraystretch{1.5}
	\centering
	\setlength{\tabcolsep}{7.0pt}
	\begin{tabular}{|c|l|c|c|}
		\hline
		\multirow{3}{*}{\shortstack[c]{LiDAR \\ specification}}  & vertical angle [deg.]   & -15 $\sim$ 15 \\
		                        								 & horizontal angle [deg.] & -180 $\sim$ 180 \\
									 	                         & max. range [m]    	   & 80 \\
		\hline
		\multirow{2}{*}{Scenes}     & train set (\# of scenes) & Town~01 $\sim$ 06 (~22,244~) \\
                                    & test set (\# of scenes)  & Town~07 \& 10 (~2,847~) \\
		\hline
	\end{tabular}
	\vspace{-0.2cm}
	\label{tab:datasets}
\end{table}

\subsection{Experimental Settings}
\label{sec:exp_setting}

We select the bilinear interpolation algorithm as a baseline approach, which computes the weights of four neighbors. 
Also, the state-of-the-art implicit method, LIIF~\cite{chen2021learning}, is evaluated to check the effectiveness of our weight prediction approach.
These methods, including ours, aim to solve Eq.~\ref{eq:upsampling}, LiDAR super-resolution without resolution constraint. 
On the other hand, LiDAR-SR~\cite{shan2020simulation} up-scales the low-resolution range image to its trained resolution only. 
Using the pixel-based LiDAR super-resolution method, we check the benefits of the implicit model. 

The experiments perform LiDAR super-resolution, which up-scales the range image from the low $16 \times 1024$ resolution to higher resolutions.
In the test, we use three test resolutions; $64 \times 1024$, $128 \times 2048$, and $256 \times 4096$.
We train a single model of each implicit network with the $128 \times 2048$ resolution data to evaluate it in the various test resolutions, including both in- and out-of-distributions.
To reconstruct range images at a specific test resolution, we make a set of query lasers matching pixels' center.
Note that it needs to train a pixel-based network at a fixed upscale factor to compare performance.
Therefore, we train and evaluate individual LiDAR-SR networks at each test resolution setting in this experiment.

We use two Tesla V100 32GB GPUs except training the LiDAR-SR network for $256 \times 4096$ resolution; it requires four GPUs. 
On the PyTorch framework, we train the prior methods with the parameters reported in their papers.
Specifically, we train these models by Adam optimizer~\cite{kingma2014adam} with an initial learning rate $10^{-4}$. The batch size is set to $16$.

The LiDAR super-resolution networks reconstruct the up-scaled range image having test resolution. We measure the mean absolute error (MAE) of all the pixels in the predicted 2-D range images.
Furthermore, we measure the performances using the 3-D points reconstructed by networks.
Since various applications use the LiDAR point cloud as raw sensor data, the reconstruction performances represent methods' usefulness.
Specifically, we measure the representation accuracy of the points with 0.1~m grid; intersection over union (IoU), precision and recall, and F1 score. 
These metrics show how well a method reconstructs LiDAR points similar to ground truth points.

\begin{table}[t]
	\caption{Quantitative comparison for LiDAR data reconstruction on CARLA dataset. 
		The bold texts represent the best performance on each metric. 
		$^{*}$Pixel-based super-resolution networks were trained to reconstruct each target resolution individually.
	}
	\renewcommand \arraystretch{1.4}
	\centering
	\setlength{\tabcolsep}{6.0pt}
	\begin{tabular}{l c c c c c}
		\hline
		\multicolumn{1}{c}{Method} & MAE & IoU & Precision & Recall & F1 \\
		\hline
	    \hline
	    \multicolumn{6}{c}{Test resolution: $64 \times 1024$} \\
	    \hline
		LiDAR-SR~\cite{shan2020simulation}$^{*}$ & 1.560 & 0.233 & 0.370 & 0.377 & 0.373 \\ \cdashline{1-6}[1pt/4pt]
		Bilinear & 2.372 & 0.202 & 0.322 & 0.328 & 0.325 \\ 
		LIIF~\cite{chen2021learning} & 1.558 & 0.258 & 0.403 & 0.409 & 0.406 \\
		Ours & \textbf{1.536} & \textbf{0.329} & \textbf{0.483} & \textbf{0.486} & \textbf{0.484} \\
		\hline
		\hline
		\multicolumn{6}{c}{Test resolution: $128 \times 2048$} \\
		\hline
		LiDAR-SR~\cite{shan2020simulation}$^{*}$ & 1.746 & 0.161 & 0.262 & 0.288 & 0.274 \\ \cdashline{1-6}[1pt/4pt]
		Bilinear & 2.591 & 0.165 & 0.268 & 0.287 & 0.277 \\ 
		LIIF~\cite{chen2021learning} & 1.714 & 0.236 & 0.372 & 0.388 & 0.379 \\
		Ours & \textbf{1.690} & \textbf{0.331} & \textbf{0.483} & \textbf{0.498} & \textbf{0.491} \\
		\hline
		\hline
		\multicolumn{6}{c}{Test resolution: $256 \times 4096$} \\
		\hline
		LiDAR-SR~\cite{shan2020simulation}$^{*}$ & \textbf{1.735} & 0.127 & 0.207 & 0.245 & 0.224 \\ \cdashline{1-6}[1pt/4pt]
		Bilinear & 2.646 & 0.163 & 0.256 & 0.303 & 0.277 \\ 
		LIIF~\cite{chen2021learning} & 1.923 & 0.158 & 0.221 & 0.356 & 0.272 \\
		Ours & 1.763 & \textbf{0.232} & \textbf{0.353} & \textbf{0.396} & \textbf{0.373} \\
		\hline
	\end{tabular}
	\vspace{-0.2cm}
	\label{tab:result_carla}
\end{table}

\begin{figure}[t]
	\centering
	\subfigure [MAE - $128 \times 2048$]  { \includegraphics[width=4.0cm]{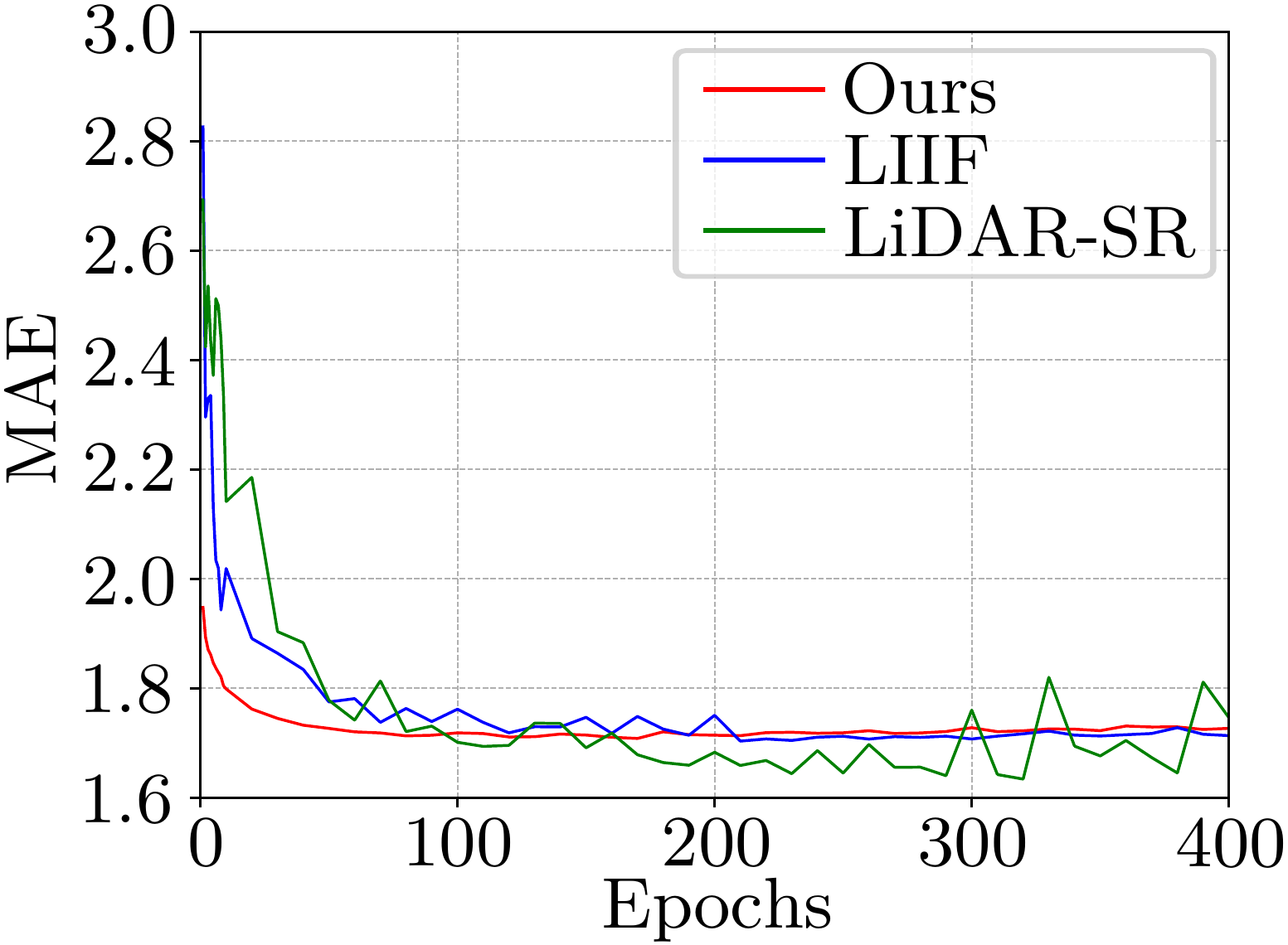} }
	\subfigure [IoU - $64 \times 1024$]   { \includegraphics[width=4.0cm]{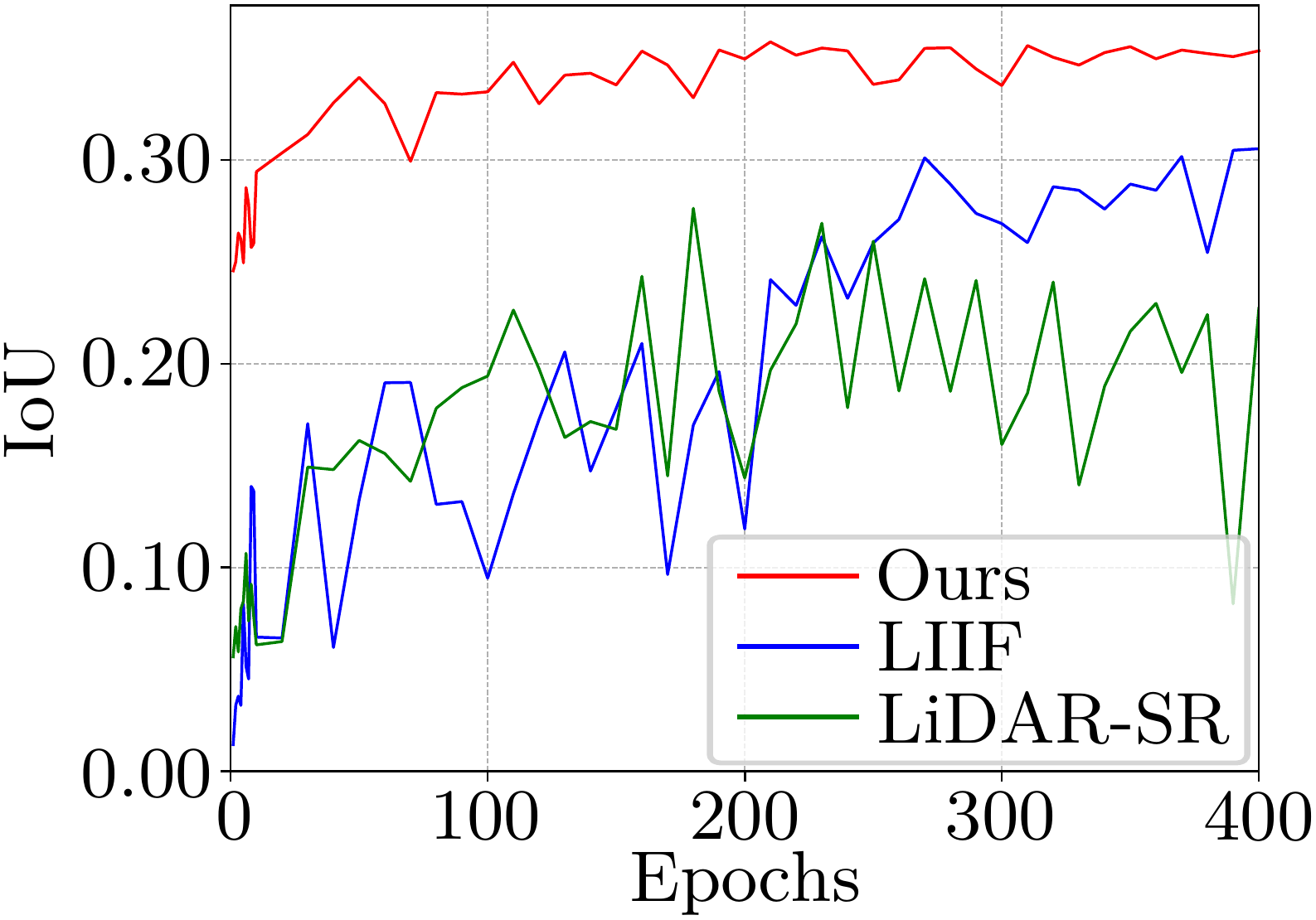} }
	\subfigure [IoU - $128 \times 2048$]  { \includegraphics[width=4.0cm]{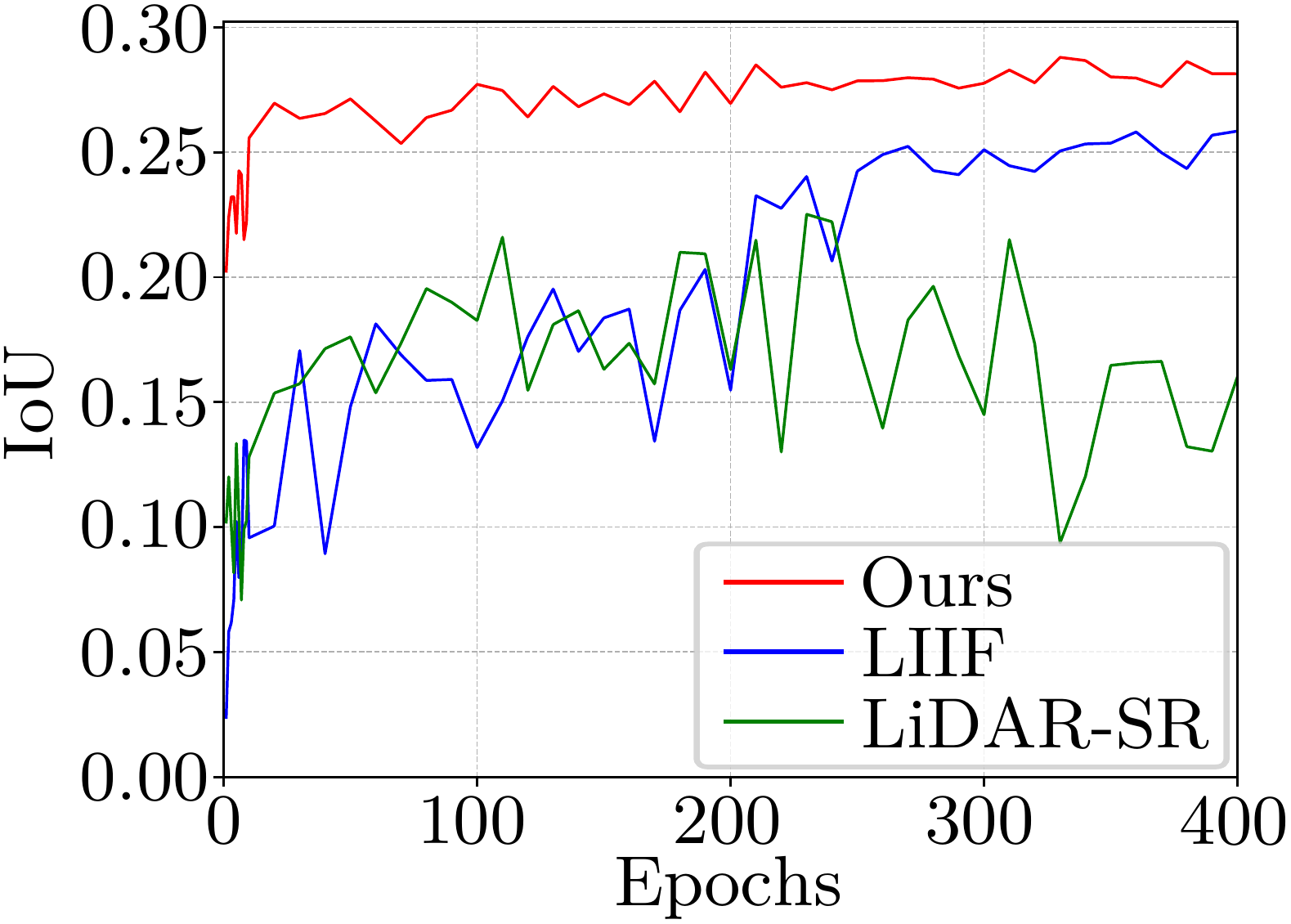} }
	\subfigure [IoU - $256 \times 4096$]  { \includegraphics[width=4.0cm]{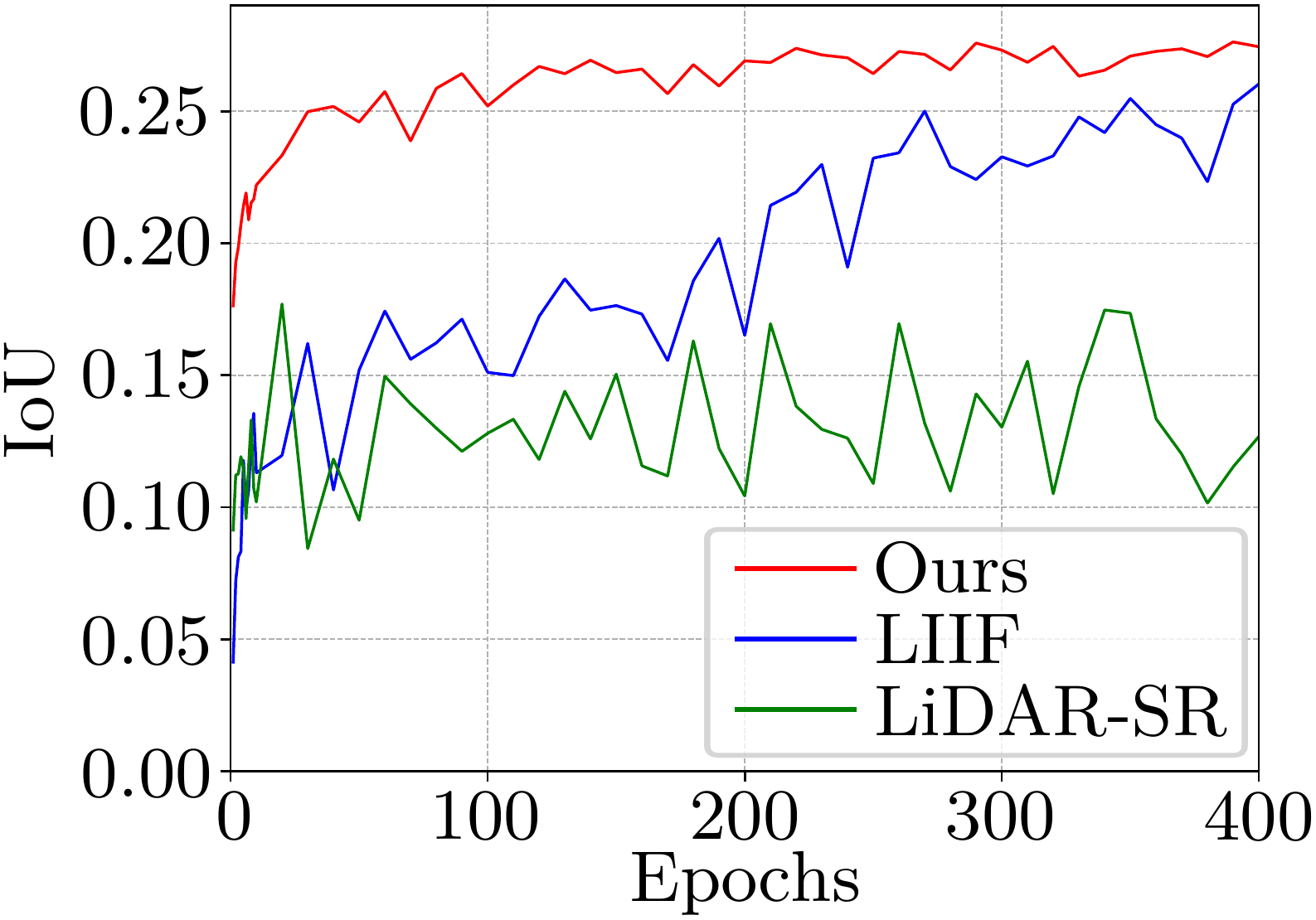} } 
	\caption{
	    Performances on the test set according to training epochs. 
	    We report the evaluation results at every 10 epochs.
	    Comparing with the other methods, ours shows the outstanding convergence speed with stable performance.
	}
	\label{fig:converence}
\end{figure}

\begin{figure*}[t]
	\centering
    \includegraphics[width=\textwidth]{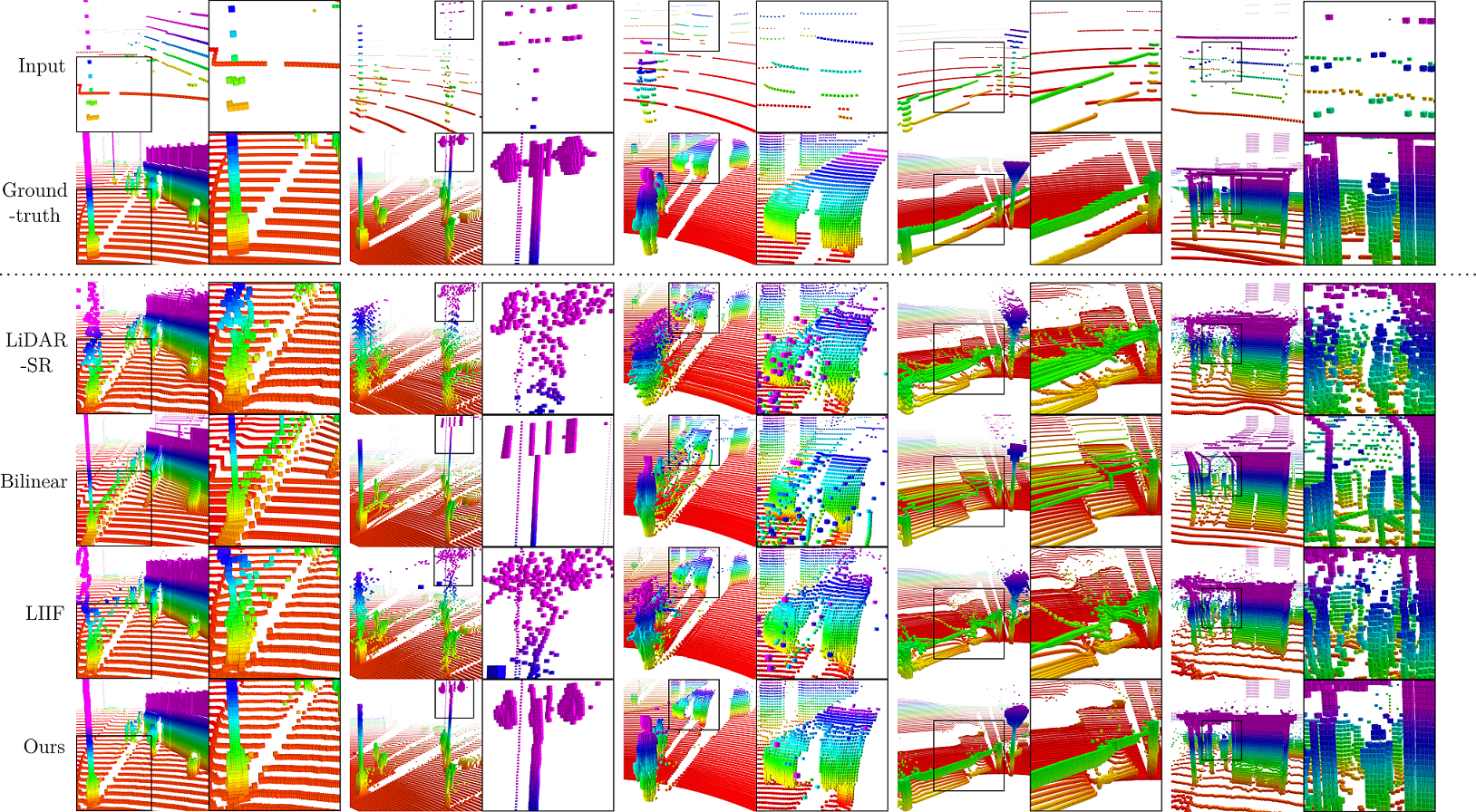}
    \vspace{-0.6cm}
	\caption{
		The qualitative results of LiDAR super-resolution via various methods. 
		The highlighted region in the black box of each left figure is shown in its right side. 
		Compared to the other methods, ours reconstructs the 3-D points robustly with much less noisy artifacts. 
		The color represents a relative height.
	}
	\vspace{-0.3cm}
	\label{fig:qualitative}
\end{figure*}

\subsection{Comparison with Prior Methods}
\label{sec:comparison}

In this section, we demonstrate the benefits of our method comparing with the prior methods. 
In summary, Table~\ref{tab:result_carla} reports the quantitative performances and Fig.~\ref{fig:qualitative} shows the qualitative results at the $128 \times 2048$ test resolution setting.

\textbf{Comparison with weight computation approach.}
When comparing the approaches estimating the interpolation weights, our method outperforms the bilinear interpolation on all the metrics.
The bilinear approach and ours utilize the same reference values from four neighbor pixels.
Nonetheless, the reported performances demonstrate that our deep network predicts the interpolation weights more robustly than the bilinear weight computation.
The deep network exploits the prior knowledge and estimates the adaptive interpolation weights based on deep features. 
As a result, ours reports up to 2.0 times improvement on the IoU evaluation metric.

\textbf{Comparison with implicit network.}
Our implicit network predicts the interpolation weights to compute detection distances of query lasers, while LIIF predicts the values (Fig.~\ref{fig:sharp_transition}-$(b)$). 
Two implicit methods are trained with the resolution data only; thus, the experiments using the various test resolutions can have different data distributions from the training dataset.
The test resolutions less than equal to the training resolution, $64 \times 1024$ and $128 \times 2048$, indicate the \emph{in-distribution} test environments. 
Otherwise, the $256 \times 4096$ resolution becomes \emph{out-of-distribution}.

Table~\ref{tab:result_carla} reports the experimental results using both in-distribution and out-of-distribution tests. 
Overall, our method outperforms the state-of-the-art implicit network in the various settings.
On the in-distribution test, ours achieves higher performance than the prior work in both 2-D range image and 3-D points reconstruction. 
In particular, our method achieves remarkable performance gains for representation accuracy of reconstructed LiDAR points, as shown in Fig.~\ref{fig:qualitative}. 
For example, our method shows 0.330 IoU performances on average of two test resolutions, while LIIF reports 0.247 IoU. 
Furthermore, we achieved significant performance improvements on the out-of-distribution test.
Our method shows outperforming 3-D points reconstruction, while reporting the meaningful improvement on the MAE metric. 
This result represents that our implicit model can cover continuous queries at an even higher resolution.

We can achieve such improvements thanks to the interpolation weights prediction instead of the values. 
Like the qualitative results in Fig.~\ref{fig:qualitative}, we observed that the value predictions of LIIF can lead undesired noisy artifacts on 3-D representation.
On the other hand, our weight prediction approach reconstructs the dense LiDAR points robustly.
In our model, the predicted weights $w_{1:4}$ determine the valuable reference values $r_{1:4}$ of the input range image via local deep features $\mathbf{z}_{1:4}'$ and self-attention mechanism.
The proposed method learns how to blend the input pixel values to fill the unmeasured information through non-linear weights. 
As a result, the approach shows the outstanding performances for LiDAR super-resolution through quantitative and qualitative results as well.

\textbf{Comparison with pixel-based super-resolution.}
Ours and LIIF are based on implicit network structures. 
On the other hand, LiDAR-SR has pixel-based convolution/deconvolutional architecture. In this analysis, we check the benefits of our method based on implicit function. 

Table~\ref{tab:result_carla} shows the our method reports the much higher performances on evaluation metrics, except slight lower performance in MAE at the $256 \times 4096$ case.
Note that since we train our implicit model with the lower resolution, $128 \times 2048$, than the test resolution, $256 \times 4096$.
On the other hand, the LiDAR-SR network was trained for the test resolution. Despite such conditions, our model shows outstanding 3-D points reconstruction with a similar MAE.

Our model shows such performance gains with a single trained network. 
The experiments using various test resolutions show that our implicit network can predict the detection distance $\hat{r}$ of query laser $\mathbf{q}$ given the sparse sensor observations, without resolution constraint. Furthermore, our network shows the robust LiDAR points reconstruction in the various test resolutions, compared to the pixel-based networks trained with each test resolution data. 

\textbf{Convergence speed.}
Our method predicts the detection distance $\hat{r}$ based on predicted interpolation weights with reference values of the input range image. 
This architecture design results in a significant convergence speed, as shown in Fig.~\ref{fig:converence}.
These graphs show that our method converges faster than other methods in various metrics and test resolutions, reporting more stable performances. 
Such fast convergence speed helps train a new model in a different environment without huge costs.

\subsection{Effectiveness of Self-Attention}
\label{sec:ablation_study}

\begin{figure}[t]
	\centering
	\includegraphics[width=\columnwidth]{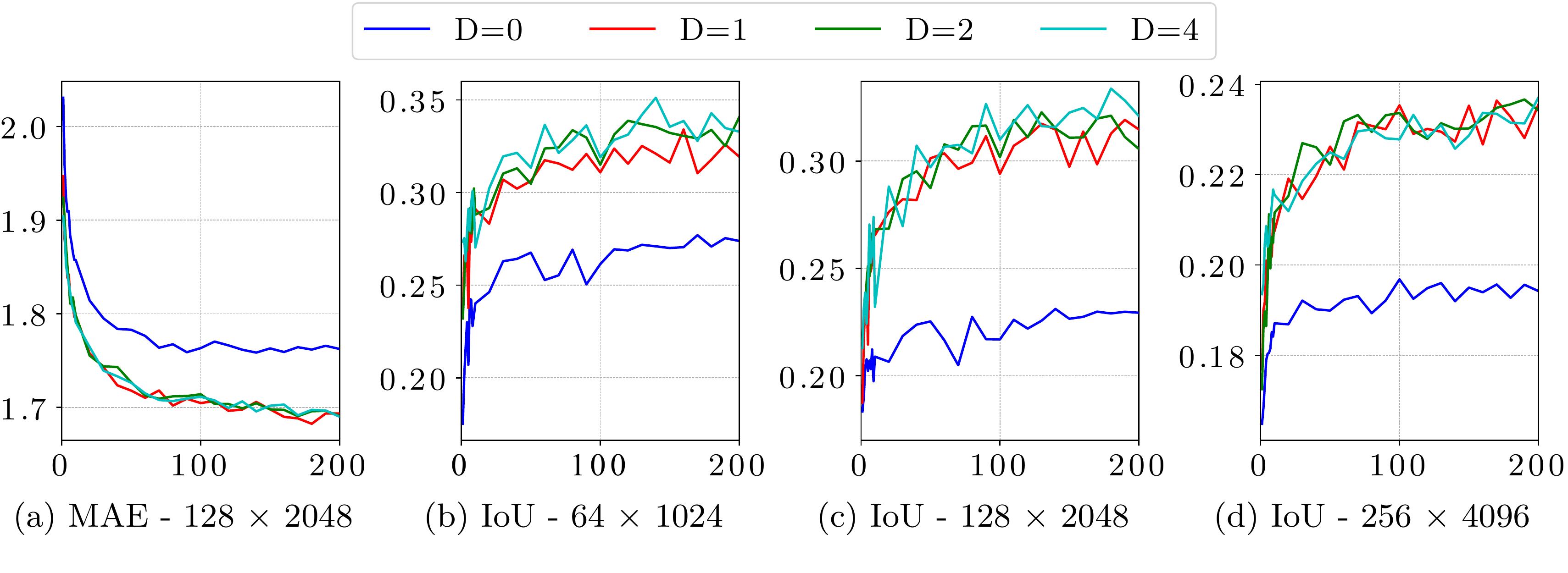}
 	\vspace{-0.8cm}
	\caption{
	    Performances of ours depending on the number of attentions, $D$.
	}
	\label{fig:ablation}
	\vspace{-0.2cm}
\end{figure}

Our network utilizes an attention mechanism to achieve performance improvement, as mentioned in Sec.~\ref{sec:weight_estimation}.
To show such benefits of attention in our model, we evaluate the performance gains over different numbers of self-attentions.
The graphs in Fig.~\ref{fig:ablation} show the experimental results at various test resolutions.
We observed remarkable performance gains over all the tests when comparing the model with and without the attention module, $D=1$ and $D=0$, respectively. 
On the other hand, applying more self-attentions showed slight performance improvement. 
These experimental results demonstrate the effectiveness of the self-attention mechanism in our model.

%% file: 4_conclusion.tex
\section{Conclusion}
\label{sec:conclusion}

We have proposed an Implicit LiDAR Network (ILN), learning an implicit function for LiDAR range image super-resolution. 
Inspired by recent work, LIIF~\cite{chen2021learning}, our network views the LiDAR image as continuous 2-D data and predicts the depth at the given query point 
by taking the depths in neighbor input pixels and interpolating them. 
However, in contrast to LIIF, learning depths (values) for the neighbor pixels and \emph{linearly} interpolates them, 
our ILN learns the \emph{weights} for the interpolation and blends the input depth values with possibly non-linear learned weights, 
which significantly improves the training speed and also reconstruction accuracy, particularly for the sharp edge areas. 
Our experiments with a novel large-scale synthetic benchmark created with CARLA simulator demonstrate the outperformance of our method compared with the previous work and the pixel value prediction network.
In our future work, we would like to test the proposed approach on various real LiDAR data and environments.